\documentclass[runningheads]{llncs}
\usepackage[T1]{fontenc}

\usepackage{graphicx}
\usepackage{adjustbox}
\usepackage{caption}
\usepackage{array}
\usepackage{subcaption}
\usepackage{booktabs}
\usepackage{caption}
\usepackage{float}
\usepackage[T1]{fontenc}
\pdfoutput=1
\usepackage[font=small, labelfont=bf, width=0.9\textwidth]{caption}
\begin{document}

\title{MRI Breast tissue segmentation using nnU-Net for biomechanical modeling}

\author{Melika Pooyan\orcidID{0009-0005-7522-4062} \and
        Hadeel Awwad\orcidID{0009-0007-4054-4546} \and
        Eloy García\orcidID{0000-0002-0587-1919} \and
        Robert Martí\orcidID{0000-0002-8080-2710}}

\authorrunning{M. Pooyan et al.}
\institute{Institute of Computer Vision and Robotics, University of Girona, Spain \\
\email{\{robert.marti\}@udg.edu}}

\maketitle         
\begin{abstract}
Integrating 2D mammography with 3D magnetic resonance imaging (MRI) is crucial for improving breast cancer diagnosis and treatment planning. However, this integration is challenging due to differences in imaging modalities and the need for precise tissue segmentation and alignment. This paper addresses these challenges by enhancing biomechanical breast models in two main aspects: improving tissue identification using nnU-Net segmentation models and evaluating finite element (FE) biomechanical solvers, specifically comparing NiftySim and FEBio. We performed a detailed six-class segmentation of breast MRI data using the nnU-Net architecture, achieving Dice Coefficients of 0.94 for fat, 0.88 for glandular tissue, and 0.87 for pectoral muscle. The overall foreground segmentation reached a mean Dice Coefficient of 0.83 through an ensemble of 2D and 3D U-Net configurations, providing a solid foundation for 3D reconstruction and biomechanical modeling. The segmented data was then used to generate detailed 3D meshes and develop biomechanical models using NiftySim and FEBio, which simulate breast tissue’s physical behaviors under compression. Our results include a comparison between NiftySim and FEBio, providing insights into the accuracy and reliability of these simulations in studying breast tissue responses under compression. The findings of this study have the potential to improve the integration of 2D and 3D imaging modalities, thereby enhancing diagnostic accuracy and treatment planning for breast cancer.
\end{abstract}

\section{INTRODUCTION}
Breast cancer is the most common cancer among women, with 1 in 8 women developing invasive breast cancer in their lifetime, highlighting the need for early and accurate diagnosis to improve patient outcomes \cite{smith2013,sung2021,zhang2023}. While traditional imaging techniques provide valuable information, they have inherent limitations. Advanced methods such as multi-modality correspondence can overcome these limitations by integrating data from different sources, resulting in a more comprehensive analysis \cite{zhang2023_predicting,tan2023}. Combining imaging techniques such as mammography and MRI provides a comprehensive view of the breast, improving diagnosis and treatment planning. Mammography detects microcalcifications but struggles with dense tissue, whereas MRI excels in soft tissue contrast and detecting invasive cancers. Integrating these modalities enhances lesion detection and characterization \cite{garcia2017}. However, differences in patient positioning during imaging, such as mammographic compression and prone positioning in MRI, present challenges in integration \cite{vanengeland2003,pereira2010,rueckert1999}. Hence, advanced image registration techniques have been proposed to align these images accurately \cite{arlinghaus2011,siegler2012}.
Finite Element Analysis (FEA) commonly plays a crucial role in these registration techniques by simulating breast tissue deformation under different conditions, aiding in accurate image registration. Patient-specific models replicating the breast’s physical properties improve the precision of diagnostic and therapeutic interventions \cite{{garcia2017},{gamage2012},{melbourne2011}}. Despite advancements, the deformable nature of breast tissue complicates image correlation across modalities and clinical contexts, affecting the diagnosis, biopsy guidance, and surgical planning \cite{garcia2017}. Biomechanical modeling offers valuable insights into breast tissue behavior, understanding disease progression, and treatment planning. However, accurately identifying different tissue types within patient-specific models derived from 3D modalities like MRI is a time-consuming and error-prone manual task \cite{garcia2018_review}. Due to its high soft-tissue contrast, MRI can discriminate between different structures in the breast and enable 3D visualization \cite{giess2014}. However, breast MRI imaging includes other organs such as the lungs, heart, pectoral muscles, and thorax. As a result, it is crucial to segment the breast region from the other organs to ensure accurate analysis in biomechanical modeling.

Recent advancements in biomechanical modeling and image segmentation have made significant strides in improving breast cancer diagnosis and treatment planning. Traditional methods primarily relied on manual segmentation, which is time-consuming and prone to errors. The advent of deep learning, particularly convolutional neural networks (CNNs) such as U-Net and its variants, has revolutionized tissue segmentation in medical imaging. Hou \cite{huo2021} achieved a Dice Coefficient of 0.87 for glandular tissue using nnU-Net \cite{isensee2021}, while Zafari \cite{zafari2019} reported a Dice Coefficient of 0.89 for pectoral muscle using U-Net. Alqaoud \cite{alqaoud2022} achieved a Dice Coefficient of 0.95 for fat using a Deep Neural Network (DNN). Despite these advances, current models often segment a limited number of tissue classes and require significant manual intervention, which can reduce their clinical utility. Finite element (FE) biomechanical solvers such as NiftySim and FEBio have been used to model the mechanical properties of breast tissue, aiding in tasks such as image registration and surgical planning. However, these models often lack detailed segmentation data, which is critical for accurately simulating tissue behavior.

This paper addresses challenges in integrating 2D and 3D imaging modalities for breast cancer diagnosis and treatment planning. The main contributions of this work include:

\begin{itemize}
\item Utilized the advanced nnU-Net framework for comprehensive segmentation of all breast tissue types in breast MRI data \cite{isensee2021}. This approach addresses limitations in existing literature, which often only segment a subset of classes of breast MRI and require additional automatic or manual pre-processing steps.
\item Conducted the first known comparative analysis of NiftySim and FEBio for biomechanical modeling of breast tissue mechanics using breast MRI images. This study provides valuable insights into their relative strengths and limitations for accurately simulating breast tissue behavior.
\end{itemize}

\section{MATERIAL AND METHODS}
The private dataset comprised 166 T1-weighted non-fat saturated Dynamic contrast-enhanced MRI (DCE-MRI) scans, including follow-ups. Acquired with a 1.5 Tesla Siemens Magnetom Vision system and a CP Breast Array coil, the scans had a typical volume size of 512×256×120 voxels, with pixel spacing from 0.625 to 0.722 mm and a slice thickness of 1.3 mm. Pre-contrast volumes were primarily used for tissue segmentation. An experienced observer manually segmented the MRI volumes into seven categories: background, fatty tissue, glandular tissue, heart, lung area, pectoral muscles, and thorax. This involved labeling every 5-10 slices, with linear interpolation filling the gaps, and more precise structures segmented at smaller intervals. Thresholding techniques were used for segmenting the background, fatty, and glandular tissues based on selected regions \cite{gubern2012}.
\begin{figure}[h]
    \centering
    \includegraphics[width=\linewidth]{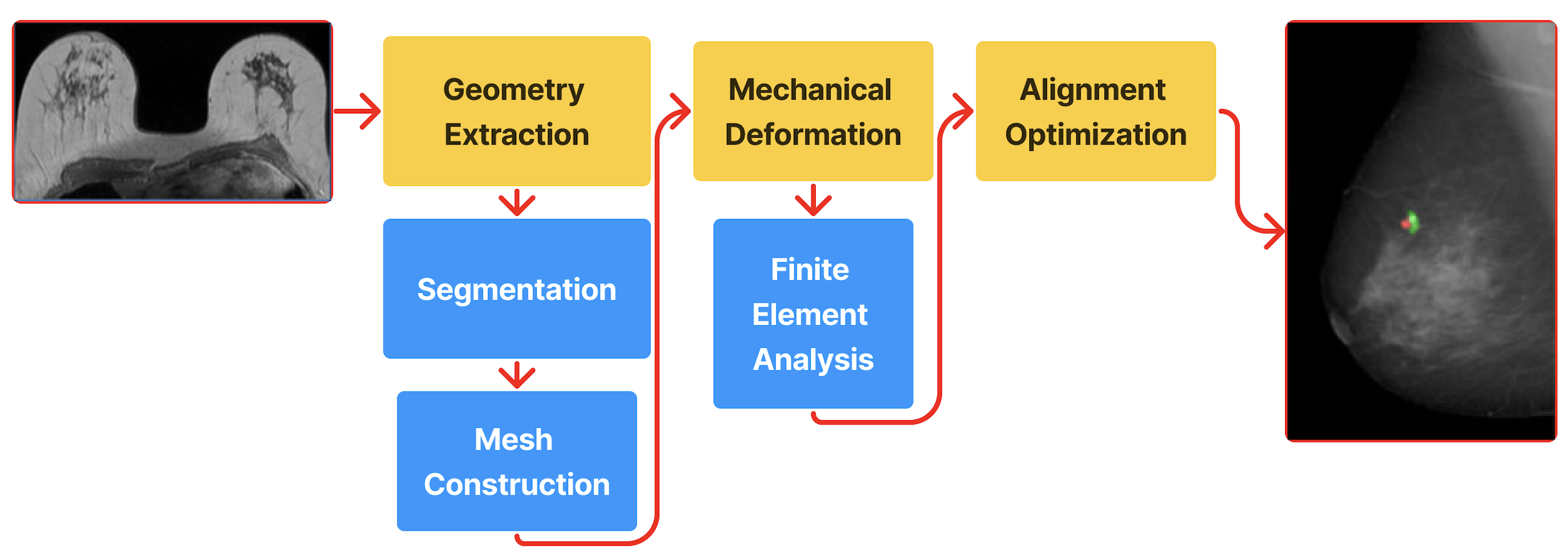}
    \caption{Overview of the steps for integrating MRI with mammography, inspired by Garcia \cite{garcia2018_review}, focusing on segmentation up to finite element analysis.}
    \label{fig:methodology}
\end{figure}
\subsection{Segmentation}
As shown in Fig. \ref{fig:methodology}, the segmentation step is a critical component of the overall process of integrating MRI with mammography. This process sets the foundation for accurate registration and biomechanical modeling. The segmentation process in this study utilized the nnU-Net framework \cite{isensee2021}, known for its high performance in medical image segmentation tasks. nnU-Net was selected due to its capability to automatically adapt its architecture to the specific dataset, thereby optimizing performance without the need for extensive manual configuration \cite{isensee2021}.
The segmentation involved several critical steps:
\begin{itemize}
    \item Data Preprocessing: MRI volumes were normalized and resampled to an isotropic voxel size to ensure uniformity across the dataset.
    \item Training Configuration: The nnU-Net architecture was configured based on the dataset's characteristics, including selecting appropriate hyperparameters, loss functions (a combination of Dice and cross-entropy loss), and optimization algorithms (stochastic gradient descent with Nesterov momentum).
    \item Model Training: Separate models were trained using both 2D and 3D U-Net configurations. The 2D U-Net processed individual slices of MRI volumes, while the 3D U-Net handled volumetric data, providing a comprehensive analysis of the tissue structures.
    \item Ensembling: The final segmentation results were obtained by ensembling the outputs from the 2D and 3D models. This involved averaging the softmax probabilities from both configurations to generate the final segmentation labels.
\end{itemize}
Details on the dataset fingerprint, which includes the dataset characteristics identified by nnU-Net such as image size, voxel spacing, and intensity distributions, as well as the hyperparameters determined based on these characteristics for both 2D and 3D networks, and the architectures of the 2D and 3D networks, are provided in the supplementary material.
\subsection{Geometry Extraction and Mesh Generation} 
Following the segmentation step in the overall process of integrating MRI with mammography, as shown in Fig. \ref{fig:methodology}, the geometry extraction and mesh generation process is the next critical step. The geometry extraction and mesh generation process begins by utilizing the segmentation results obtained from the nnU-Net framework \cite{isensee2021}. The initial step involves isolating the breast region from the MRI volumes, excluding non-breast tissues. This isolation is achieved by applying a pre-obtained breast region mask from Gubern-Mérida \cite{gubern2012}, which effectively segments the image background, leaving only the volumes of interest, such as fat and glandular tissue. The sternum serves as a reference point to ensure accurate segmentation. Following the segmentation, the isolated breast volume, including its internal fat and glandular tissues, is resampled to isotropic voxels of 1 mm³. Although nnU-Net automatically resamples data based on the median image spacing of the dataset, this additional resampling after segmentation ensures consistency and better mesh quality. The volume mesh is then generated using pygalmesh \cite{schlomer2021}, a Python interface for CGAL's meshing tools \cite{cgal2024}. This tool is capable of generating both 2D and 3D meshes. The element count in these meshes varies between 50,000 and 500,000, depending on the volume of the breast, which helps minimize errors during finite element simulations \cite{garcia2017}.
\subsection{Finite Element Analysis: Simulating Compression}
\begin{figure}[h]
    \centering
    \includegraphics[width=0.65\linewidth]{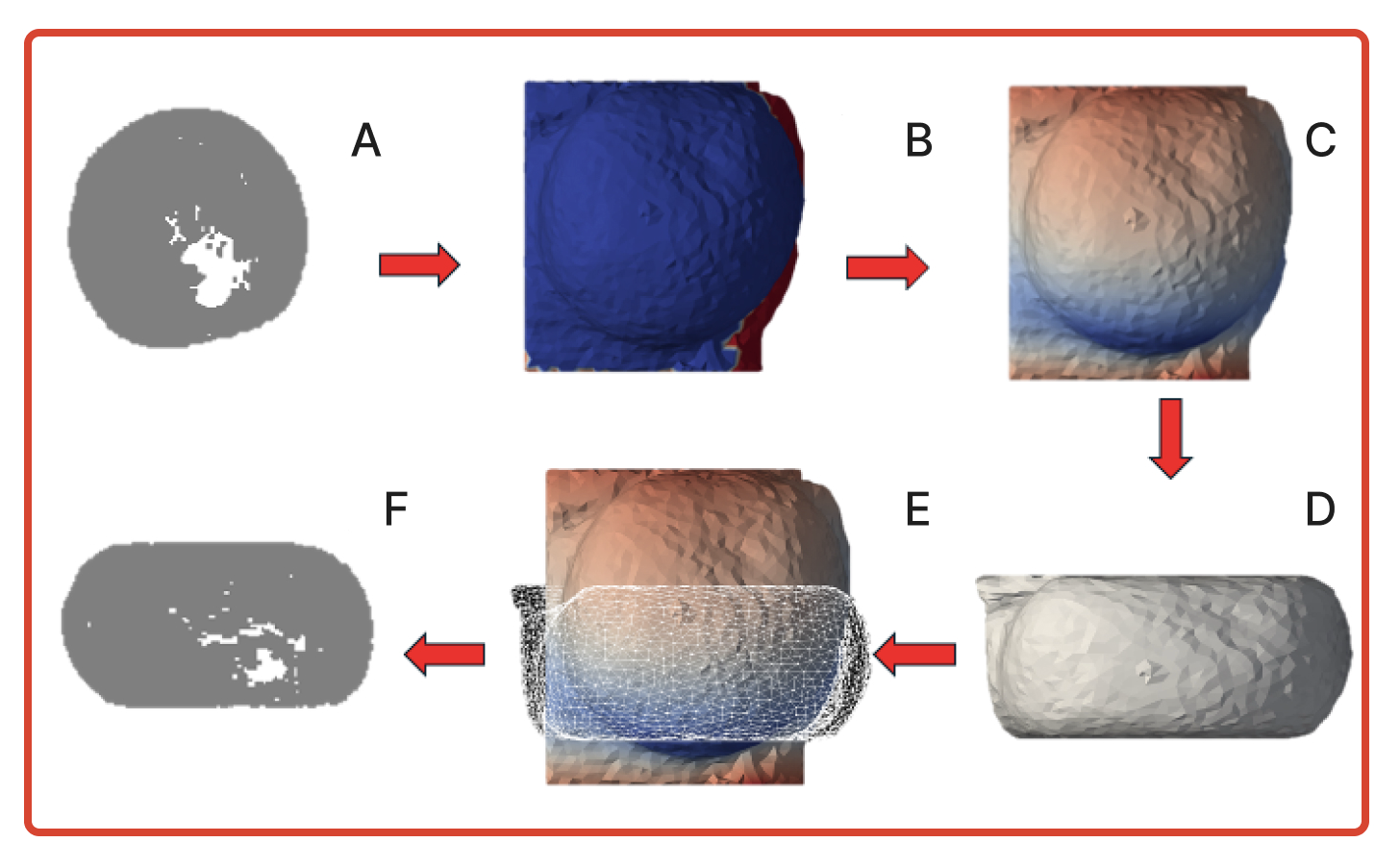}
    \caption{Process of compression: (A) Segmentation Map, (B) Generated Mesh, (C) NiftySim Displacement, (D) Compressed Map, (E) Wireframe overlay comparing pre- and post-compression maps, (F) Final Compressed Map.}
    \label{fig:nifty}
\end{figure}

Finite Element Analysis (FEA), which is the next step in the pipeline, is essential for simulating the mechanical behavior of breast tissue under conditions like mammography compression. FEA models were constructed using segmented MRI data, incorporating mechanical properties to simulate deformation and stress distribution accurately. NiftySim \cite{garcia2020} and FEBio \cite{maas2012} are open-source software for biomechanical simulations of soft tissues. They support properties like position, and orientation of the patient, to adapt the registration process to the patient-specific conditions. Moreover, the initial parameters of the elastic materials were set based on literature values reported in the work of Garcia \cite{garcia2020}, specifically Young’s modulus (4.46 kPa for fatty tissue, 15.1 kPa for glandular tissue) and Poisson’s ratio (0.45 to 0.499). Both tools generate uncompressed and compressed breast models, suitable for detailed analysis under different conditions. NiftySim's efficiency in handling large-scale simulations made it ideal for this study \cite{garcia2020}. FEBio offers advanced features for simulating complex tissues and incorporates sophisticated material models and boundary conditions. It has been used to simulate breast compression using high-resolution CT data, handling detailed anatomical models and complex tissue interactions \cite{hsu2011}. In this study, FEBio validated and compared NiftySim's results. Using FEBio, breast tissue's response to mechanical forces were analyzed, further validating NiftySim's results. The compression process is similar for both tools, as illustrated for NiftySim in Fig. ~\ref{fig:nifty}. 
\section{EXPERIMENTAL RESULTS}
\subsection{EVALUATION METRICS}
The performance of the segmentation and biomechanical modeling processes was evaluated using two critical metrics: the Dice Coefficient and breast volume (BV) measurements. These metrics are essential for assessing the accuracy of breast tissue deformation under compression in our study. Firstly, the Dice Coefficient was utilized to measure segmentation accuracy by quantifying the overlap between the predicted and ground truth labels. Additionally, it was employed to assess the accuracy of biomechanical modeling using NiftySim and FEBio by comparing compressed and uncompressed segmentation maps, focusing on the center of mass for fat and glandular tissues. A high Dice score close to 1 indicates that the tissues did not deform significantly under compression, suggesting the need for further analysis to ensure accurate simulation. To complement the Dice Coefficient, breast volume changes were analyzed to evaluate the model's ability to simulate realistic tissue behavior under compression. Ideally, the breast volume should remain constant, indicating no tissue loss. However, due to inherent imperfections in simulations, a smaller reduction in breast volume is preferable, indicating better compression with minimal tissue loss. This metric is crucial for understanding the extent of tissue deformation and loss during compression. The study by Garcia \cite{garcia2020} supports the use of breast volume changes as an evaluation metric, highlighting its relevance in biomechanical modeling. For a comprehensive statistical analysis, the mean and standard deviation (SD) of the deviations were examined between the analyzed cases. These metrics allow for assessing the consistency and reliability of the segmentation and biomechanical modeling processes, offering insights into the overall performance and robustness of the models.
\subsection{Segmentation results}
\begin{figure}[h]
    \centering
    \includegraphics[width=1\linewidth]{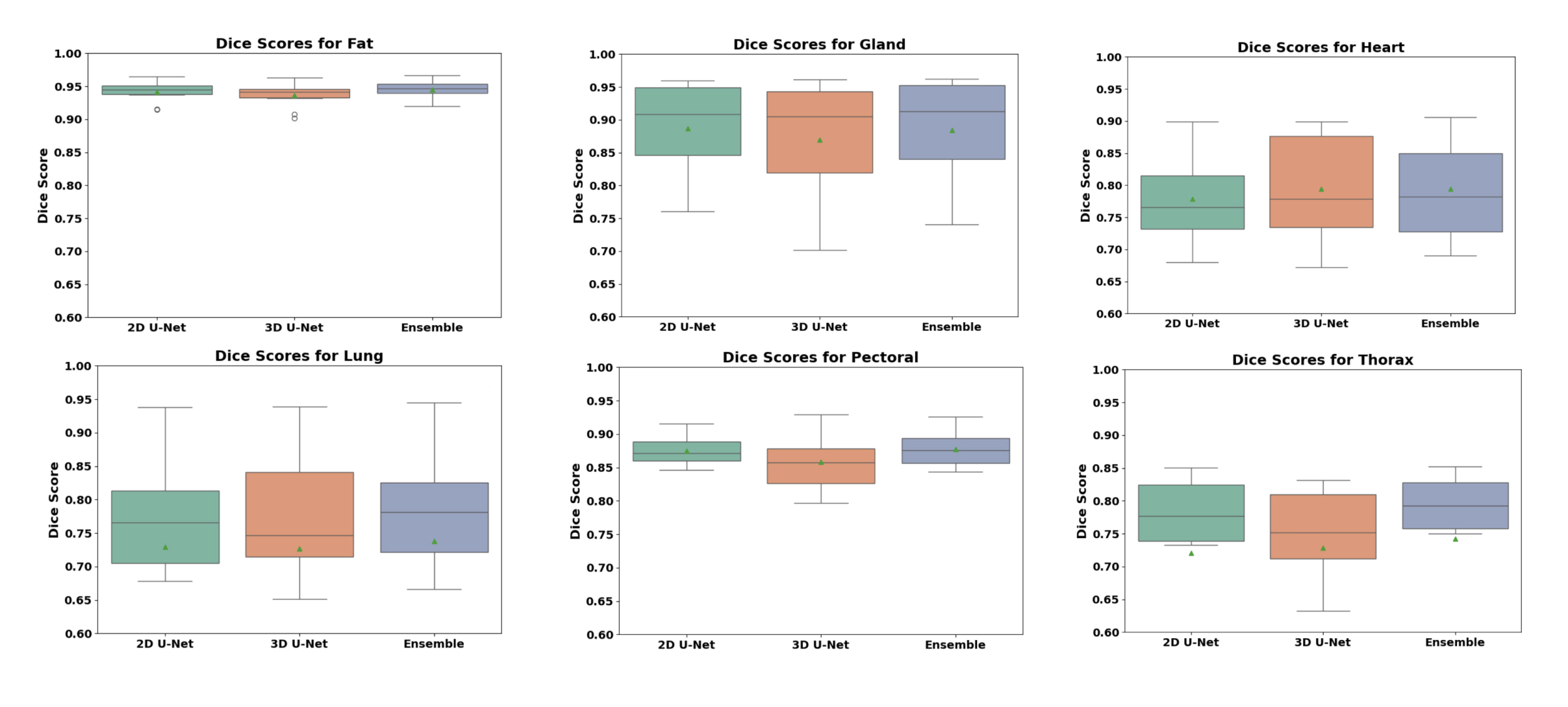}
    \caption{Dice Coefficients for six tissue types segmented by 2D U-Net, 3D U-Net, and their ensemble, shown from a scale of 0.60 as the Dice Coefficients for all classes were above this value.}
    \label{fig:boxplot}
\end{figure}

\begin{figure}[h]
    \centering
    \includegraphics[width=\linewidth]{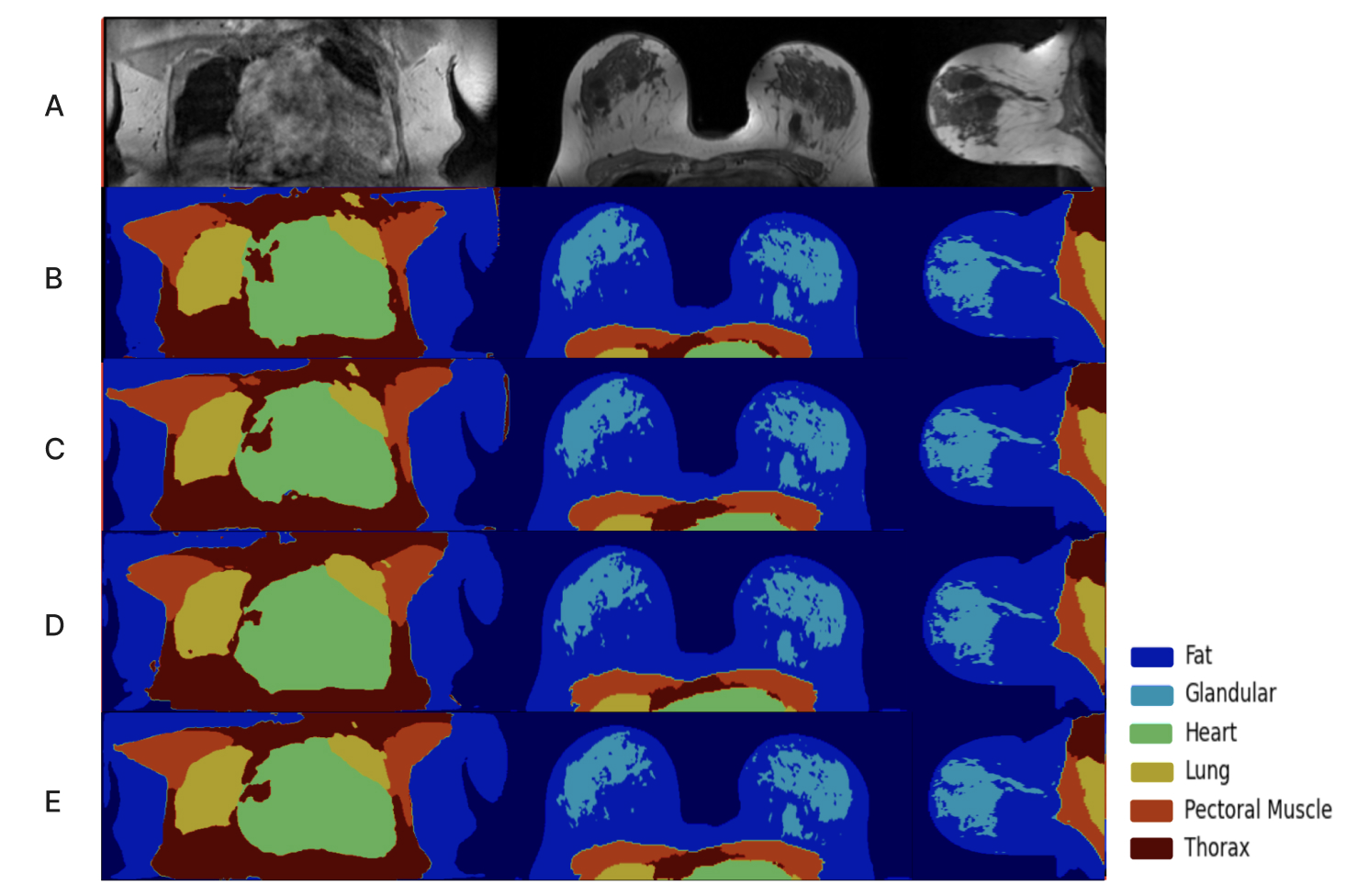}
    \caption{Qualitative segmentation results for six tissue types (Fat, Glandular, Heart, Lung, Pectoral Muscle, and Thorax) using MRI data. A: Original MRI images. B: Ground truth segmentation. C: 2D U-Net results. D: 3D U-Net results. E: Ensemble method results. The ensemble method (E) shows the most consistent and accurate segmentation, closely matching the ground truth (B).}
    \label{fig:qseg}
\end{figure}

The nnU-Net framework demonstrated high performance in segmenting breast tissues and organs, including in breast MRI data. The quantitative results, summarized in Table~\ref{tab:comparison_methods}, show robust segmentation accuracy across different tissue types. The Dice Coefficients indicate that the framework effectively captures the details of breast tissues, comparable to state-of-the-art methods in the literature. Additionally, the mean and standard deviation (SD) values provide an overview of the average segmentation performance and the variability across different tissues, indicating consistent performance by the nnU-Net framework. The boxplots demonstrate the Dice coefficients for six tissue types segmented using 2D U-Net, and 3D U-Net, and their ensemble. The ensemble method generally shows higher or similar median Dice Coefficients compared to the individual 2D and 3D U-Nets, especially for fat and pectoral tissues. The narrower interquartile ranges for the ensemble method in tissues like fat and pectoral suggest more consistent performance, while individual methods show more variability (Fig. ~\ref{fig:boxplot}). Detailed boxplots, particularly highlighting the high Dice coefficients for the fat class, are included in the supplementary material. Moreover, visual assessments confirmed the accuracy of the segmentation, accurately delineating tissue boundaries even in challenging regions. These segmentation results provide a strong foundation for subsequent biomechanical modeling and analysis, as illustrated in Fig. ~\ref{fig:qseg}.

\begin{table}[h]
    \caption{Comparison of nnU-Net results with State-of-the-Art \cite{huo2021,zafari2019,alqaoud2022}.}
    \label{tab:comparison_methods}
    \centering
    \renewcommand{\arraystretch}{1.3} 
    \resizebox{\textwidth}{!}{
    \begin{tabular}{c|c|c|c|c|c|c|c}
        \hline
        \textbf{Methods} & \textbf{Fat} & \textbf{Glandular} & \textbf{Heart} & \textbf{Lung} & \textbf{Pectoral} & \textbf{Thorax} & \textbf{Mean $\pm$ SD} \\
        \hline
        2D-UNet & \textbf{0.94} & \textbf{0.88} & 0.77 & 0.72 & \textbf{0.87} & 0.72 & 0.82 $\pm$ 0.14 \\
        \hline
        3D-UNet & 0.93 & 0.86 & \textbf{0.79} & 0.72 & 0.85 & 0.72 & 0.81 $\pm$ 0.13 \\
        \hline
        Ensemble & \textbf{0.94} & \textbf{0.88} & \textbf{0.79} & \textbf{0.73} & \textbf{0.87} & \textbf{0.74} & \textbf{0.83 $\pm$ 0.13} \\
        \hline
        State of the Art & 0.95 \cite{alqaoud2022} & 0.87 \cite{huo2021} & - & - & 0.89 \cite{zafari2019} & - & - \\
        \hline
    \end{tabular}
    }
\end{table}

\subsection{Biomechnical modeling results}
A subset of 10 cases was chosen to obtain their biomechanical models. Out of these, 4 cases were successfully compressed, while the rest were not, potentially due to issues with the mesh or segmentation affecting the biomechanical models. The biomechanical modeling results, summarized in Table 2, show that NiftySim consistently outperformed FEBio in modeling accuracy and breast volume preservation. NiftySim achieved higher Dice Coefficients for both fat (0.78 to 0.91) and glandular tissues (0.19 to 0.31) compared to FEBio's lower values for fat (0.59 to 0.72) and glandular tissues (0.14 to 0.28). Despite concerns about higher Dice Coefficients after compression, NiftySim showed less breast volume loss (1.52\% to 1.94\%) compared to FEBio (3.22\% to 4.30\%), indicating better preservation of anatomical integrity and more accurate tissue deformation modeling. Additionally, mean and standard deviation (SD) values for these measurements are included in Table 2.
\begin{table}[h]
    \caption{Dice Coefficients, BVs of two FEA Methods for 4 Cases}
    \label{tab:FEAdices}
    \centering
    \renewcommand{\arraystretch}{1.2} 
    \begin{tabular}{c|c|c|c|c}
        \toprule
        \textbf{Cases} & \textbf{FEA} & \textbf{Fat} & \textbf{Gland} & \textbf{BV} \\
        \midrule
        Case 1 & NiftySim & 0.91 & 0.22 & 1.52\% \\
        Case 1 & FEBio & 0.69 & 0.20 & 3.22\% \\
        \midrule
        Case 2 & NiftySim & 0.78 & 0.31 & 1.63\% \\
        Case 2 & FEBio & 0.59 & 0.28 & 4.11\% \\
        \midrule
        Case 3 & NiftySim & 0.85 & 0.20 & 1.87\% \\
        Case 3 & FEBio & 0.72 & 0.19 & 4.30\% \\
        \midrule
        Case 4 & NiftySim & 0.89 & 0.19 & 1.94\% \\
        Case 4 & FEBio & 0.65 & 0.14 & 4.18\% \\
        \midrule
        \textbf{Mean $\pm$ SD} & \textbf{NiftySim} & 0.85 $\pm$ 0.05 & 0.23 $\pm$ 0.05 & - \\
        \textbf{Mean $\pm$ SD} & \textbf{FEBio} & 0.66 $\pm$ 0.05 & 0.20 $\pm$ 0.05 & - \\
        \bottomrule
    \end{tabular}
\end{table}

\section{DISCUSSION AND CONCLUSIONS}
In this work, we presented a comprehensive approach for six-class segmentation of breast MRI data using the nnU-Net framework, followed by detailed biomechanical modeling with NiftySim and FEBio. Our study aims to compare and analyze the performance of these tools in segmenting and modeling breast tissues, thereby providing insights into their respective strengths and limitations.
The nnU-Net framework demonstrated high Dice Coefficients and precise tissue boundaries, effectively segmenting all breast tissue types. In the comparative analysis, NiftySim generally outperformed FEBio in biomechanical modeling, achieving expected Dice Coefficients for fat and glandular tissues with less volume loss. This indicates that NiftySim may provide a superior simulation of tissue biomechanics under compression, maintaining anatomical integrity during simulations. Accurate biomechanical models facilitate the correlation of breast structures across imaging modalities, support CAD algorithms and needle biopsy procedures, and help radiologists evaluate suspicious areas over time. Despite these advancements, only 4 out of the 10 cases analyzed were successfully compressed. This limited success rate may be due to segmentation issues, mesh quality, or other complexities in finite element analysis, highlighting the need for further research and improvement in these areas.

In conclusion, while the nnU-Net framework effectively segments breast tissue types in MRI data and NiftySim shows promise in biomechanical modeling, the current success rate indicates significant areas for improvement. Challenges such as segmentation accuracy, mesh quality, and the complexity of finite element analysis need to be addressed to enhance the robustness and reliability of biomechanical simulations. Recognizing both the strengths and the areas needing improvement, this work lays the foundation for future advancements in breast tissue segmentation and biomechanical modeling. Future work should focus on refining these aspects to improve simulation success rates, better support personalized treatment planning, and ultimately improve outcomes for patients undergoing breast cancer diagnosis and treatment.

\begin{credits}
\subsubsection{\ackname} 
This work has been partially funded by the Erasmus+: Erasmus Mundus Joint Master’s Degree (EMJMD) scholarship (2022–2024), with project reference 610600-EPP-1-2019-1-ES-EPPKA1-JMD-MOB and the project VICTORIA, “PID2021-123390OB-C21” from the Ministerio de Ciencia e Innovación of Spain.

\subsubsection{Disclosure of Interests.}
The authors have no competing interests to declare that are relevant to the content of this article.

\end{credits}


\bibliographystyle{splncs04}

\section{Supplementary materials}
\subsection{Segmentation}
The dataset fingerprint played a crucial role in the segmentation process. This fingerprint includes specific properties identified by nnU-Net, such as image size and voxel spacing, which were used to determine the optimal hyperparameters for each configuration. By leveraging these dataset-specific characteristics, nnU-Net was able to adapt its architecture and training process to maximize performance and accuracy \cite{isensee2021}. This tailored approach ensured that the segmentation process was both precise and efficient, laying a strong foundation for subsequent biomechanical modeling (Tabel \ref{tab:networkconfigs}).

\begin{figure}[h]
    \centering
    \includegraphics[width=1\linewidth]{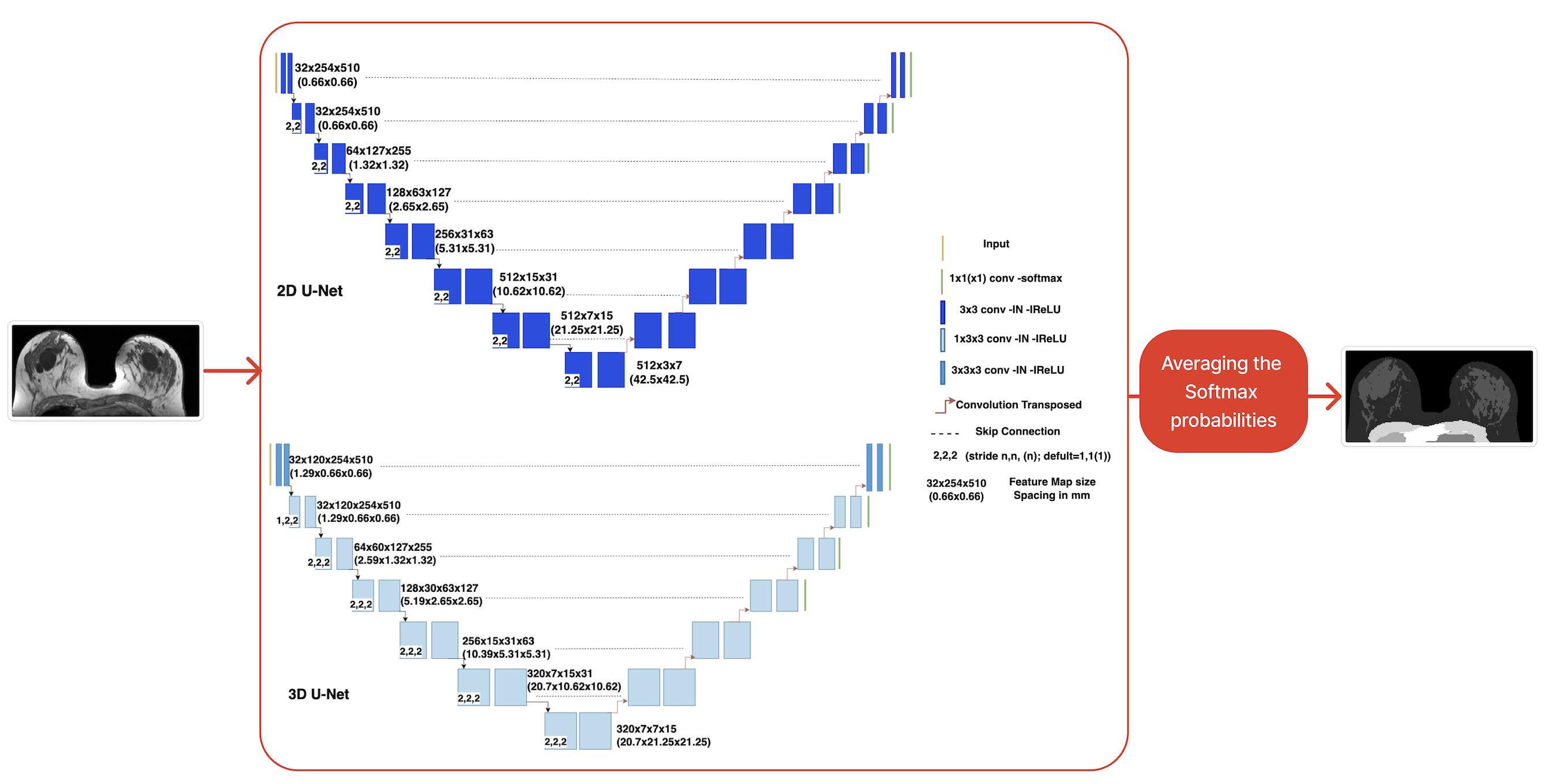}
    \caption{Network architectures generated by nnU-Net for the dataset \cite{isensee2021}.}
    \label{fig:nnunetframe}
\end{figure}

The architecture generated by nnU-Net, which was utilized in this study, automatically adapts to the dataset's characteristics, optimizing the network for improved performance without extensive manual configuration (Fig. \ref{fig:nnunetframe}).

\begin{table}[h]
    \centering
    \caption{Dataset fingerprint and hyperparameters for 2D and 3D U-Net.}
    \label{tab:networkconfigs}
    \Large
    \renewcommand{\arraystretch}{1.9} 
    \begin{tabular}{>{\centering\arraybackslash}m{0.33\linewidth} >{\centering\arraybackslash}m{0.33\linewidth} >{\centering\arraybackslash}m{0.33\linewidth}}
        \begin{adjustbox}{max width=\linewidth}
        \begin{tabular}{|l|c|}
            \hline
            \multicolumn{2}{|c|}{\textbf{Dataset}} \\ \hline
            Median image size & 120x254x510 \\ \hline
            Median image spacing & 1.29x0.66x0.66mm \\ \hline
            Normalization & Z-score \\ \hline
        \end{tabular}
        \end{adjustbox}
        &
        \begin{adjustbox}{max width=\linewidth}
        \begin{tabular}{|l|c|}
            \hline
            \multicolumn{2}{|c|}{\textbf{2D-UNet}} \\ \hline
            Target Spacing & NAx254x510 \\ \hline
            Median Shape @ Target Spacing & NAx0.66x0.66mm \\ \hline
            Patch Size & 256x512 \\ \hline
            Batch Size & 24 \\ \hline
        \end{tabular}
        \end{adjustbox}
        &
        \begin{adjustbox}{max width=\linewidth}
        \begin{tabular}{|l|c|}
            \hline
            \multicolumn{2}{|c|}{\textbf{3D-fullres UNet}} \\ \hline
            Target Spacing & 120x254x510 \\ \hline
            Median Shape @ Target Spacing & 1.29x0.66x0.66mm \\ \hline
            Patch Size & 64x128x288 \\ \hline
            Batch Size & 2 \\ \hline
        \end{tabular}
        \end{adjustbox}
    \end{tabular}
\end{table}

\subsection{Segmentation results}
The detailed boxplots provide a thorough look at the Dice coefficients for six tissue types segmented using 2D U-Net, and 3D U-Net, and their ensemble. The fat class, being a predominant tissue type, achieved significantly higher Dice coefficients compared to the other tissue classes across all models.

\begin{figure}[h]
    \centering
    \includegraphics[width=1\linewidth]{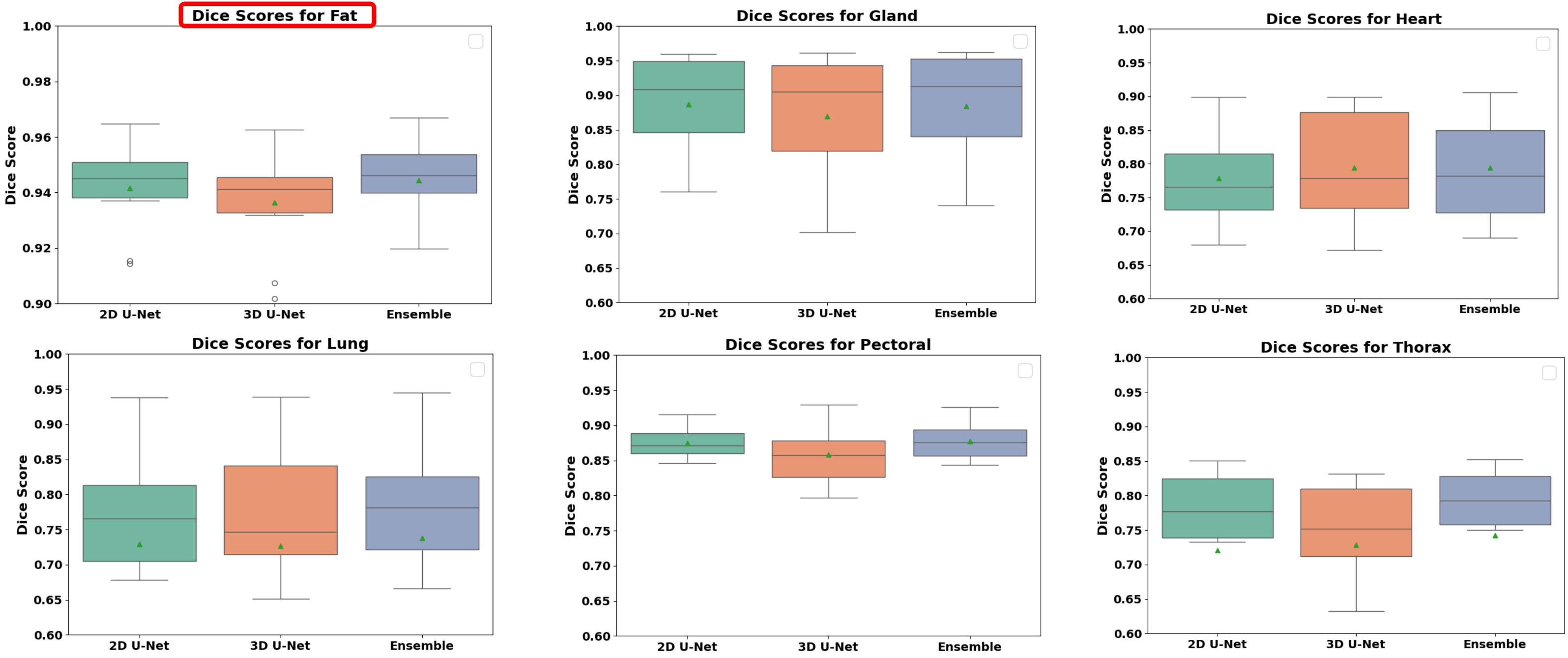}
    \caption{Dice scores for six tissue types segmented by 2D U-Net, 3D U-Net, and their ensemble, with the fat class shown on a scale starting from 0.90.}
    \label{fig:boxplotfat}
\end{figure}

The boxplots reveal that the ensemble method generally produces the most consistent results, with the fat class exhibiting Dice coefficients consistently above 0.90. This superior performance underscores the robustness of the segmentation for fat tissues, which is critical given its majority presence in breast tissue (Fig. \ref{fig:boxplotfat}).

\end{document}